\pdfoutput=1
\documentclass[10pt,twocolumn,letterpaper]{article}

\usepackage{cvpr}
\usepackage{times}
\usepackage{epsfig}
\usepackage[breaklinks=true,bookmarks=false]{hyperref}
\usepackage{capt-of}
\usepackage{color}
\usepackage{algpseudocode}
\usepackage{algorithm}

\usepackage{graphics, graphicx}
\usepackage{subfigure}
\usepackage{epstopdf}
\usepackage{float}
\usepackage{multirow}
\usepackage{cite}
\usepackage{latexsym,bm,amsmath,amssymb}
\usepackage{enumerate}
\usepackage[noblocks]{authblk}
\usepackage{flushend}
\usepackage{dcolumn,longtable,hhline,colortbl}
\usepackage{latexsym,bm,amsmath,amssymb}
\usepackage{color}

\newcommand{\PreserveBackslash}[1]{\let\temp=\\#1\let\\=\temp}
\newcolumntype{C}[1]{>{\PreserveBackslash\centering}p{#1}}
\newcolumntype{R}[1]{>{\PreserveBackslash\raggedleft}p{#1}}
\newcolumntype{L}[1]{>{\PreserveBackslash\raggedright}p{#1}}

\newcommand{\tabincell}[2]{\begin{tabular}{@{}#1@{}}#2\end{tabular}}

\cvprfinalcopy 

\begin{document}

\title{Dating Ancient Paintings of Mogao Grottoes Using Deeply Learnt Visual Codes}


\author[1]{Qingquan Li}
\author[2]{Qin Zou\thanks{Corresponding: qzou@whu.edu.cn}}
\author[3]{De Ma}
\author[2]{Qian Wang}
\author[4]{Song Wang}
\affil[1]{\small{Shenzhen Key Laboratory of Spatial Smart Sensing and Service, Shenzhen University, P.R.~China}}
\affil[2]{School of Computer Science, Wuhan University, P.R.~China}
\affil[3]{Dunhuang Research Academia, Dunhuang, P.R.~China}
\affil[4]{Department of Computer Science and Engineering, University of South Carolina, USA}
\maketitle
\begin{abstract}
Cultural heritage is the asset of all the peoples of the world. The preservation and inheritance of cultural heritage is conducive to the progress of human civilization. In northwestern China, there is a world heritage site -- Mogao Grottoes -- that has a plenty of mural paintings showing the historical cultures of ancient China. To study these historical cultures, one critical procedure is to date the mural paintings, i.e., determining the era when they were created. Until now, most mural paintings at Mogao Grottoes have been dated by directly referring to the mural texts or historical documents. However, some are still left with creation-era undetermined due to the lack of reference materials. Considering that the drawing style of mural paintings was changing along the history and the drawing style can be learned and quantified through painting data, we formulate the problem of mural-painting dating into a problem of drawing-style classification. In fact, drawing styles can be expressed not only in color or curvature, but also in some unknown forms -- the forms that have not been observed. To this end, besides sophisticated color and shape descriptors, a deep convolution neural network is designed to encode the implicit drawing styles. 3860 mural paintings collected from 194 different grottoes with determined creation-era labels are used to train the classification model and build the dating method. In experiments, the proposed dating method is applied to seven mural paintings which were previously dated with controversies, and the exciting new dating results are approved by the Dunhuang expert.
\end{abstract}

\section{Introduction}
Mogao Grottoes, known as the Thousand Buddha Caves, is located in the western end of the Hexi Corridor along the ancient Silk Road in Dunhuang, China. It was initially built in 366 AD in the period of the Five Huns Sixteen Regime. Afterwards, with large constructions in the Northern Dynasties, Sui, Tang, Wu Dai, Song, Xixia and Yuan Dynasties, a huge scale has been formed which includes 735 caves, 45,000m$^2$ mural paintings, and 2,415 argillaceous painted sculptures. The Mogao Grottoes is known as the world's largest and the richest Buddhist art place, and was listed as a World Heritage Site in 1987.

Culture heritage is a precious asset of humanity. For the purpose of culture-heritage preservation and historical study, each grotto is assigned an official era label by experts at Dunhuang Research Academy{\footnote{http://en.dha.ac.cn/}}. This task is commonly known as dating the grotto. To date a grotto, there are mainly three sources to acquire information. The first source is the mural texts directly written on the walls of the grotto, which stand as a part of the mural painting in the grotto and directly describe the creation event of the grotto or the mural paintings. The second source is the historical documents and manuscripts that record information of the construction event. And the third source is the drawing style of the mural paintings, under the assumption that the drawing styles in the same era or dynasty are similar. In situations lacking the first and the second sources, the third source will play a vital role in dating the grottoes and the paintings.

In ancient China, Mogao Grottoes not only appears as a religious place, but also serves as a political symbol. According to historical documents, mural paintings in Mogao Grottoes were drawn by official painters, who should not be certificated to draw paintings until they reach a master's level through a strict official training. In this way, mural paintings can be drawn in similar drawing styles and strictly reflect the faith and spirit of the central government. When a dynasty was replaced by another one, the drawing style would consequently change. For example, ladies with slim figure were very popular in Sui Dynasty, while fat figures were appreciated in the succeeded Tang Dynasty. As a result, fairy ladies and human ladies in Tang's mural paintings were fatter than that in Sui's. Meanwhile, red colors and long smooth curves were more often used in Tang's mural paintings than in Sui's. Generally, the overall drawing style of one dynasty is different from that of another, and hence a mural painting commonly attaches a label indicating the era of its creation.

\begin{figure*}[t!]
\centering
\begin{minipage}[b]{1\linewidth}
  \centering
  \hspace{2mm} {\includegraphics[width=16.2cm]{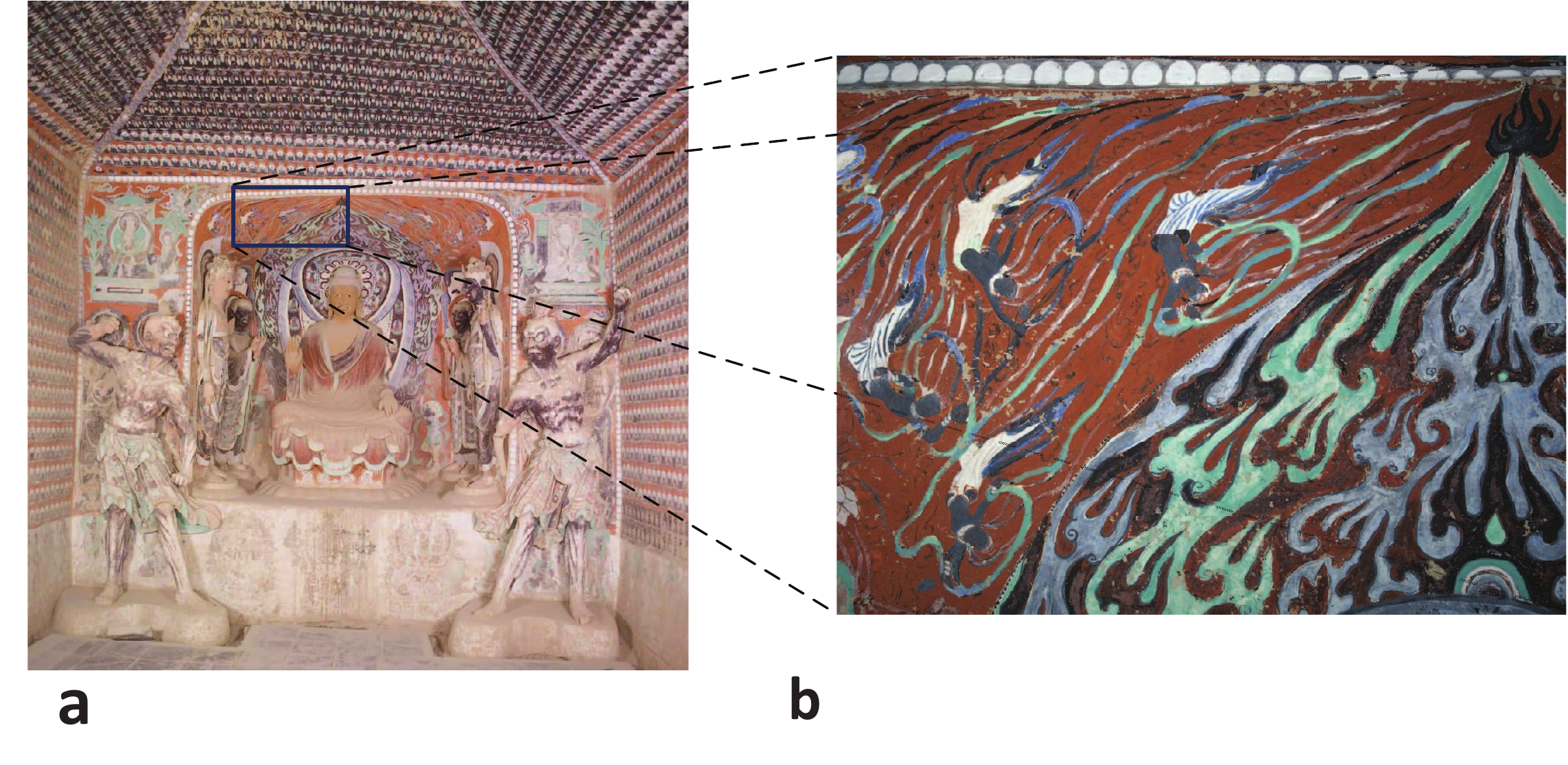}}
  \medskip
\end{minipage}

\begin{minipage}[b]{0.45\linewidth}
  \centering
  \centerline{\includegraphics[width=5.98cm]{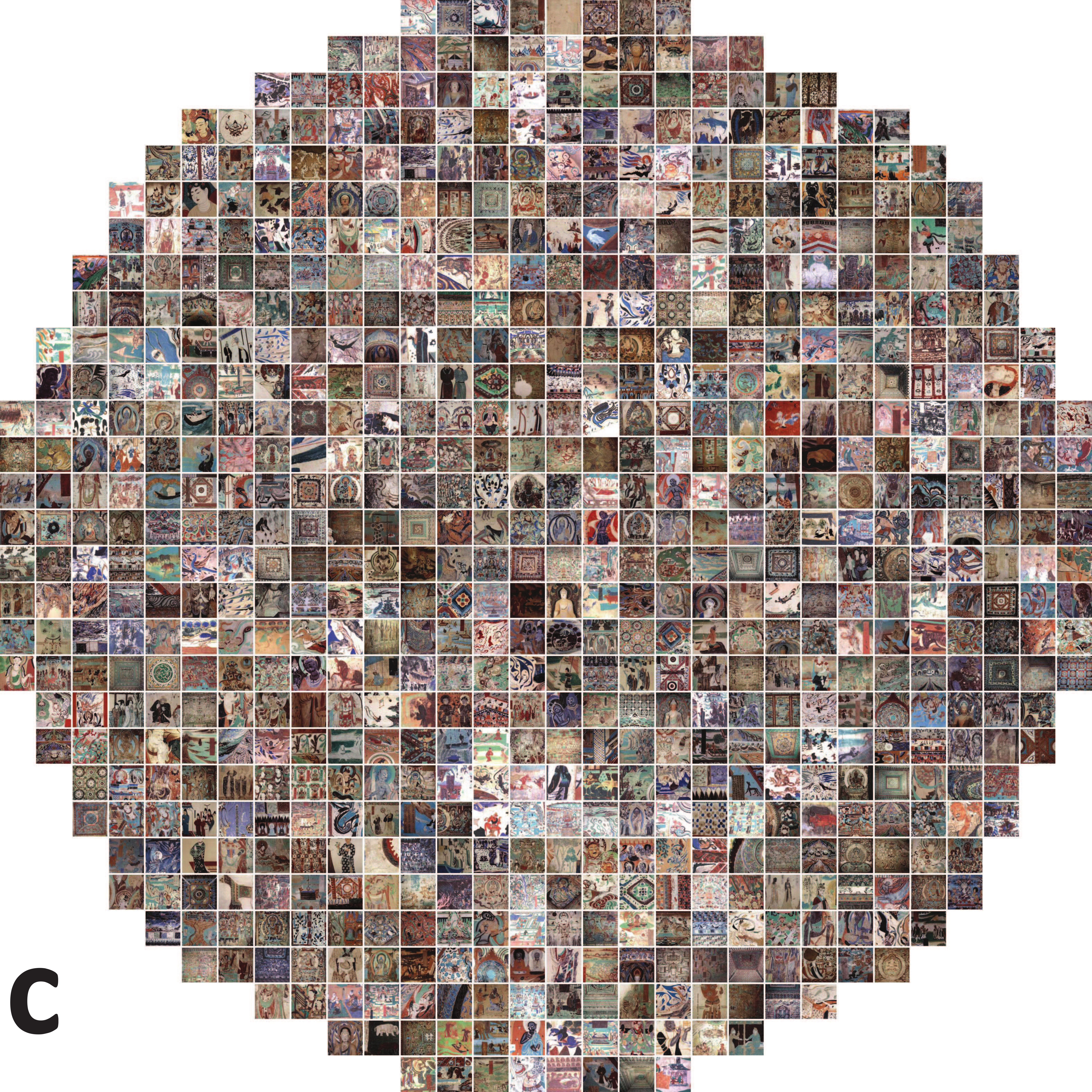}}
  \medskip
\end{minipage}
\hspace{3mm}
\begin{minipage}[b]{0.50\linewidth}
  \centering
  \centerline{\includegraphics[width=9.3cm]{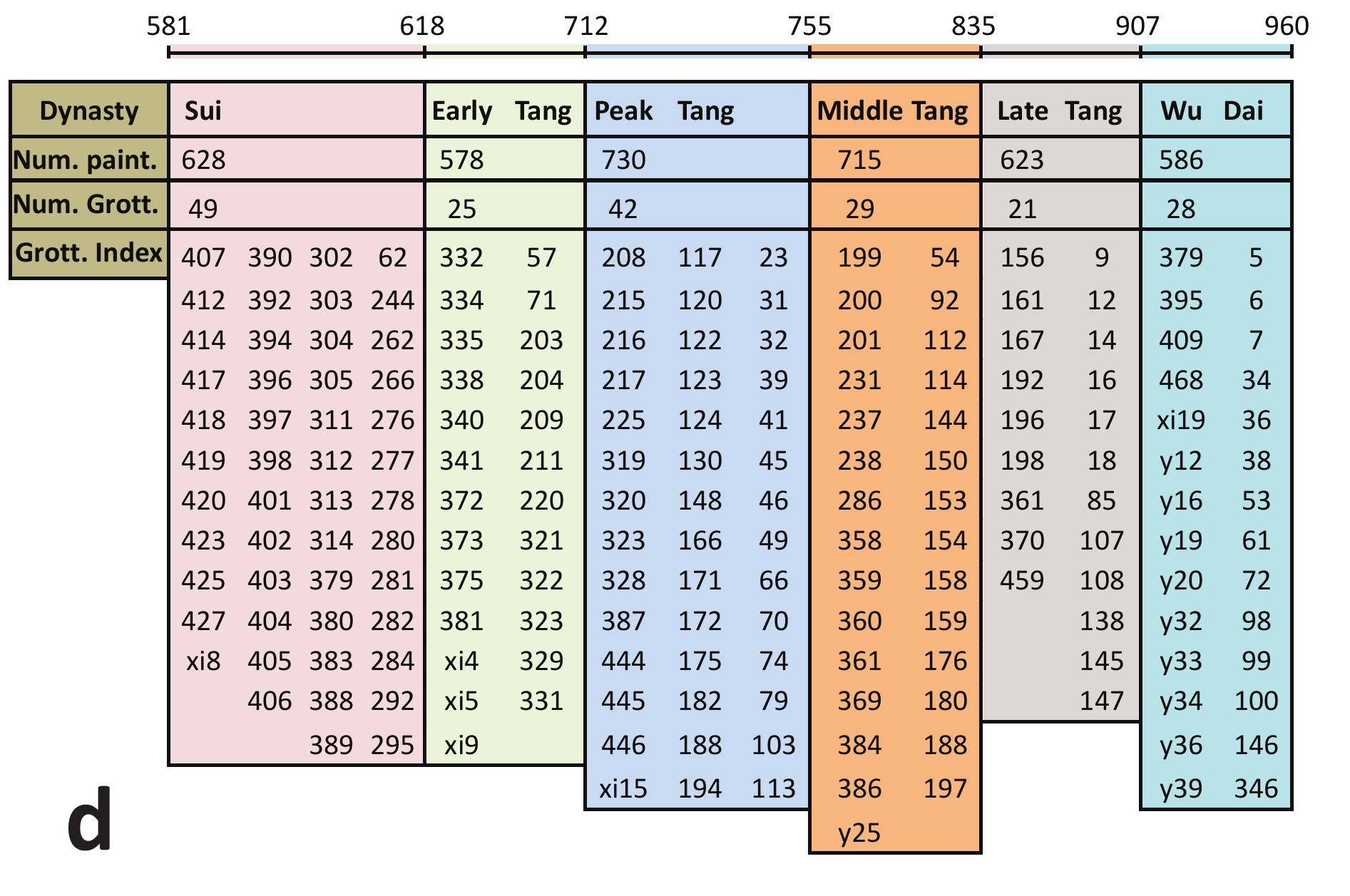}}
  \medskip
\end{minipage}\vspace{2mm}
\caption{(a) shows a part of the grotto $\texttt{\#}$206 and (b) shows a mural painting. (c) and (d) show the detail information of the DunHuang-E6 dataset. (c) Sample painting images of the DunHuang-E6. (d) The distribution of the paintings in DunHuang-E6, as regarding to the dynasty when the paintings were drawn, the number of paintings from each dynasty, the indices of the grottoes where the paintings were drawn. Note that, the marks on the top stick in (d) display the start year and the end year of each dynasty.} \label{fig:dunhuang6}
\end{figure*}

Although all grottoes and mural paintings at Mogao Grottoes have been officially dated and assigned era labels to declare their history of creation, there does exist some grottoes that have been dated in controversy. The controversy arises owing to the fact that experts sometimes cannot reach an agreement on determining the drawing style of the mural paintings, especially when mural texts and historical documents do not exist or contain any information of the creation history of the grotto or mural paintings. In such cases, the painting artists will visually examine the drawing style and compare with other era-determined paintings to predict the era of creation. However, there's no guarantee that they will reach an agreement on the drawing style or the creation era. For example, the grotto $\texttt{\#}$205 and the grotto $\texttt{\#}$206 are two typical grottoes that have neither mural texts nor historical document to identify the era of their creation. For grotto $\texttt{\#}$205, two main drawing styles were observed and declared by Dunhuang artists, which were the Peak Tang's and the Middle Tang's, as shown in Fig.~\ref{fig:seven700-7}(a)-(c). Experts guess that, the grotto $\texttt{\#}$205 was originally constructed in Early Tang or Peak Tang, but the construction was handicapped by wars of Anshi Rebellion for about 50 years, and then the construction was succeeded in Middle Tang. However, until now there's no scientific or quantitative study to support these declarations. For grotto $\texttt{\#}$206, it is located at the region of the Tang's grottoes, but it holds mural paintings of the Sui's style as pointed out by Dunhuang artists. A part of grotto $\texttt{\#}$206 and a typical mural painting have been shown in Fig.~\ref{fig:dunhuang6}(a) and (b). Generally, grottoes of the same dynasty would have been caved in the same region. To persuade people to believe that these paintings are truly in Sui's style, scientific validations are also required. The above discussions indicate that, a scientific and quantitative method to identify the style of mural paintings is strongly desired.

Computer-aided painting analysis has attracted wide attention, and in recent years has been a hot research topic. In early research, fractal techniques was employed to analyze Pollock's drip paintings and has uncovered the mysteries in drawing a drip painting~\cite{taylor1999fractal}, which pioneered the way to use computational means to analyze the painting arts. Wavelet-based methods were also employed for painting analysis. For example, the first- and higher-order wavelet statistics were used for painting art authentication~\cite{lyu2004pnas}, and Gabor wavelet filters were utilized to analyze the brush strokes of Vincent van Gogh's paintings~\cite{johnson2008image}. Meanwhile, many other advanced technologies such as sparse coding and empirical mode decomposition were employed to quantify the drawing styles, and were applied to identify the artists such as Pieter Bruegel the Elder and Dutch master Rembrandt van Rijn~\cite{hughes2010quantification,hughes2012quantitative,hughes2013sp}. As the big-data era comes, large-scale set of paintings are possible to be collected, and some general rules in painting along a historical period can be modeled and analyzed, e.g., difference in color usage between classical paintings and photographs~\cite{kim2014large}. The non-linear matrix completion was also used to model the painting style, and it successfully recognize the emotions from abstract paintings~\cite{alameda2016recognizing}. In summary, the computational and statistical algorithms built on math have been more and more used for analysis of the painting arts~\cite{ornes2015science}.

In this study, we intend to discover the creation era of mural paintings from grottoes $\texttt{\#}$205 and $\texttt{\#}$206 using computer vision and machine learning techniques. Under the assumption that drawing styles are different from dynasty to dynasty and the drawing style can be learned from painting data, the creation-era prediction problem is mapped into a drawing-style classification problem. Specifically, a method built on computer vision and machine learning techniques is proposed to quantify the drawing styles of mural paintings. In this method, a number of visual codes are extracted from a set of training mural paintings by using handicraft feature descriptors and discriminative learning algorithms, where the resulting visual codes are supposed to bring information of the drawing styles and can be used to quantify them.

Considering that the mural paintings at Mogao Grottoes have fruitful colors and some drawing styles can potentially be described by color collocations, we use a a set of advanced color descriptors to extract the color visual codes in the mural paintings. For the curves, i.e., painting strokes, which constitute an another important form of drawing style, we employ a sophisticated shape descriptor to discover the shape visual codes in the mural paintings. Besides the color and curve, the drawing styles in mural paintings of Mogao Grottoes may display in other forms such as the layout of religious scene, the scale of instruments, the central Buddha in worship, or some others that have not been observed yet, we construct a deep convolution neural network (DCNN) to encode the hidden visual codes for drawing-style description. A learning machine is then applied to process these visual codes and build a model for drawing-style classification for the paintings collected from different eras. Finally, the drawing style of a new mural painting outside the training set can be identified by using the pre-trained classification model, and the corresponding era of creation can be predicted.

\section{Materials and Methods}

\subsection{Datasets}\label{sec:data}
\ In this section, two datasets of mural paintings are introduced at first. Then the procedure in processing the painting data is described. At last, experimental settings are given.

{\it DunHuang-E6{\footnote{https://sites.google.com/site/qinzoucn/}}.} It is a dataset containing 3860 mural paintings which were painted in six continuous dynasties spanning from Sui dynasty to Wu Dai.  The 3860 painting images were collected from 194 grottoes of Mogao Grottoes, Dunhuang, China. Based on the era label assigned by Dunhuang research academia, the paintings in DunHuang-E6 are divided into six categories according to their era labels, i.e., Sui, Early Tang, Peak Tang, Middle Tang, Late Tang and Wu Dai. Detailed information of DunHuang-E6 has been given in Fig.~\ref{fig:dunhuang6}(d). The images share the same size of 400$\times$400. To obtain the classification model, there are 3000 images for training, 700 images for testing, and 160 images for validation.

{\it DunHuang-P7.} It contains seven mural paintings for creation-era prediction. Among the seven paintings, six are collected from grotto $\texttt{\#}$205 and one is collected from grotto $\texttt{\#}$206, as shown in Fig.~\ref{fig:seven700-7}. For Grotto $\texttt{\#}$205, the artists declare that two main different drawing styles exist among the mural paintings, where one is the Peak-Tang's drawing style on the south wall, e.g. Fig.~\ref{fig:seven700-7}(a), and the other is the Middle-Tang's drawing style on the west wall, e.g., Fig.~\ref{fig:seven700-7}(b) and Fig.~\ref{fig:seven700-7}(c). Scientific and quantitative evaluations are required to support the declarations. For Grotto $\texttt{\#}$206, the paintings have to be clarified if they are in Sui's or Early Tang's drawing style. All the seven images has been resized by changing their shorter sides to 600 pixels.

\begin{figure*}[t!]
    \centering
    \includegraphics[width=0.9\linewidth]{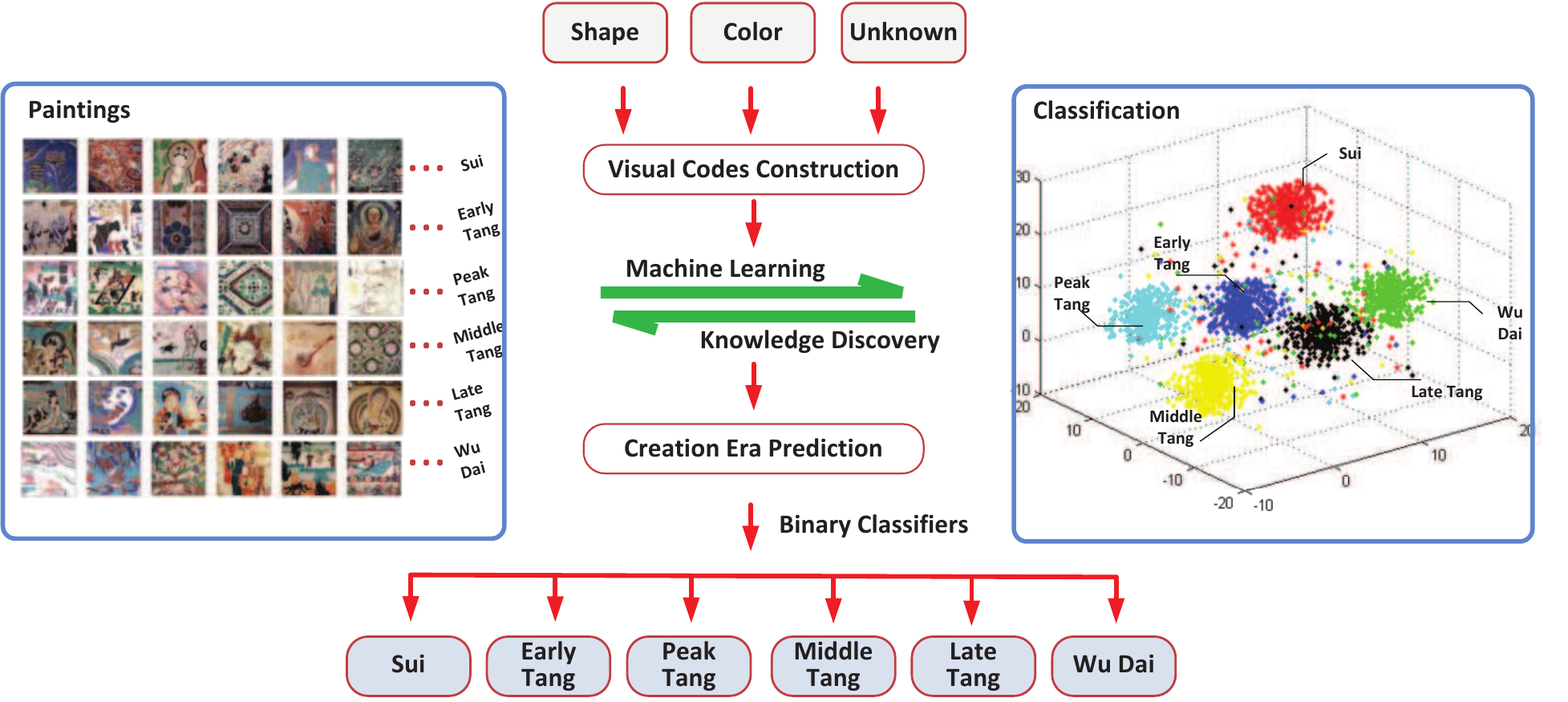}
    \caption{The system overview.}
    \label{fig:system}
\end{figure*}

\subsection{Procedure} \label{sec:method}
\ As discussed in the introduction, the creation-era prediction problem is formulated as a painting-style classification problem. The system overview is illustrated by Fig.~\ref{fig:system}. Given a set of painting images which were created in $N$ eras (or dynasties), i.e., $E_1, E_2, ..., E_N$, we construct a dataset containing $N$ classes by putting images of the same creation era into one class. We then extract visual codes and encode the drawing style in each painting image, based on which we classify the paintings into different eras using a supervised learning machine. In return, the trained classification model will provide knowledge to predict the creation-era of a new-coming painting which was painted in one of the $N$ eras.

In our case, paintings which were dated in controversy were drawn in-between the Sui dynasty and Wu Dai dynasty. Hence, we construct the dataset by collecting mural paintings from Sui to Wu Dai from Mogao Grottoes, which formed a six-class dataset for painting style classification, i.e., the Sui, Early Tang, Peak Tang, Middle Tang, Late Tang and Wu Dai. In training the classification model, shape descriptors extracted by SIFT and encoded by Improved Fisher Vector (IFV{\scriptsize{sift}}) are employed to describe drawing styles carried by curves, and color descriptors extracted by Color Names (CN) and regional color co-occurrence (RCC) descriptor are used to describe drawing styles shown in colors. To describe drawing style shown in `unknown' forms, i.e., the style forms that have not been observed, we train a DCNN on the dataset and use it as a feature extractor to describe the implicit visual codes of drawing styles.

The following will first introduce the shape-, color-, and DCNN-based drawing style description, and then present a classification-based voting strategy for dating new paintings.

\subsection{Shape-based drawing style description}\label{sec:shape}

\ The shape of the painting stroke plays as an import role in representing the drawing style of a painting. In visual perception, the shape is observed on the edges in the painting image, and the edges responds to salient gradients. In this subsection, we employ a gradient-based feature descriptor - SIFT~\cite{SIFT} as a local shape descriptor, and the IFV~\cite{perronnin2010eccv} as a shape-feature encoding method.

{\it SIFT}- Scale-Invariant Feature Transform descriptor~\cite{SIFT}. It consists of two steps: key-point detection and feature
descriptor calculation. Let $I(x,y)$ be the input image,
$L(x,y,\sigma_c)$, $c=1,2,\cdots, K$, be a sequence of smoothed
images of $I$, by convolving  $I$ with 2D Gaussian filters as
\begin{equation}\label{eq:sift-b}
L(x,y,\sigma_c)=\frac{1}{2\pi
\sigma_c^2}exp(-\frac{x^2+y^2}{2\sigma_c^2})*I(x,y),
\end{equation}
where $\sigma_c$, $c=1,2,\cdots, K$ is a sequence of monotonically
increased scale parameters. Then the difference of Gaussian (DoG)
image $D(x,y,\sigma_c)$ at scale $c$ can be computed as
\begin{equation}\label{eq:sift-a}
D(x,y,\sigma_c)=L(x,y,\sigma_{c+1})-L(x,y,\sigma_c).
\end{equation}
By stacking multiple DoG images, we actually obtain a 3D image (two
spatial dimension and one scale dimension). We then check each pixel
at each scale in its 3$\times$3$\times$3 neighborhood. If it is a
local extremum, we take it as a key
point~\cite{SIFT}.

At each key point, SIFT
descriptor is computed as the local appearance feature. For each key
point $(x,y)$, gradient magnitude $g(x,y)$ and orientation
$\theta(x,y)$  is calculated at the scale where the key point is
detected,
\begin{equation}\label{eq:sift-1}
g(x, y)=\sqrt{f_x(x,y)^2+f_y(x,y)^2},
\end{equation}
\begin{equation}\label{eq:sift-2}
\theta (x,y)=tan^{-1}\frac{f_y(x,y)}{f_x(x,y)},
\end{equation}
where $f_x(x,y)$ and $f_y(x,y)$ are calculated by
\begin{equation}\label{eq:sift-3}
\left\{
\begin{array}{rl}
f_x(x,y)=L(x+1,y, \sigma_c)-L(x-1,y, \sigma_c),\\
f_y(x,y)=L(x,y+1,\sigma_c)-L(x,y-1,\sigma_c),
\end{array}
\right.
\end{equation}
with $c$ being the scale in which the key point $(x,y)$ is detected.
Then, a weighted histogram of 36 directions is
constructed using the gradient magnitude and orientation in the
region around the key point, and the peak that is 80\% or more of
the maximum value of the histogram is selected to be the principal
orientation of the key point. After rotating the region around the
key points to the principal orientation, the region is divided into
blocks of 4$\times$4, and the histogram of eight directions is
computed at each block. Thus, a 4$\times$4$\times$8=128 element
feature vector is finally calculated as the SIFT descriptor at each
key point.

{\it IFV}- Improved Fisher Vector (IFV)~\cite{perronnin2010eccv} is an improved version of Fisher Vector (FV)~\cite{perronnin2007fisher}, which is a feature-encoding method. It quantifies a set of local image features into a global feature vector for image representation and visual classification. The FV can be derived as a special, approximate, and improved case of the general Fisher Kernel framework.

Let $I$=$(x_1, ..., x_N)$ be a set of $D$-dimensional feature vectors (e.g. SIFT descriptors) extracted from an image, then the distribution of all the feature vectors extracted from a set of images can be fitted by means of Gaussian mixture models (GMM), with a number of $K$ Gaussian models. Let $(\omega_k, \mu_k, \Psi_k)$, $k=1, ..., K$, be the weight, mean value (vector), covariance matrix of the Gaussian model $k$. The GMM associates each vector $x_i$ to a model $k$ in the mixture with a strength given by the posterior probability:
{
\begin{equation}\label{eq:fv-1}
q_{ik} = \frac{{exp}[-\frac{1}{2}(x_i-\mu_k)^{\top}\Psi_k^{-1}(x_i-\mu_k)]}{\sum_{t=1}^K {exp}[-\frac{1}{2}(x_i-\mu_t)^{\top}\Psi_t^{-1}(x_i-\mu_t)]}.
\end{equation}}
For each mode $k$, consider the mean and covariance deviation vectors
\begin{equation}\label{eq:fv-2}
u_{jk} = \frac{1}{N\sqrt{\omega_k}}\sum_{i=1}^N q_{ik}\frac{x_{ji}-\mu_{jk}}{\sigma_{jk}}.
\end{equation}
\begin{equation}\label{eq:fv-3}
v_{jk} = \frac{1}{N\sqrt{2\omega_k}}\sum_{i=1}^N q_{ik}[(\frac{x_{ji}-\mu_{jk}}{\sigma_{jk}})^2-1].
\end{equation}
where $\sigma_{jk}$ =$|\Psi_{jk}|^{1/2}$ and $j$=1,2, ...,$D$ spans the vector dimensions. Suppose $\textbf{u}_k$ = [$u_{1k},u_{2k},..., u_{Dk}$], and $\textbf{v}_k$ = [$v_{1k},v_{2k},..., v_{Dk}$], then the FV of image $I$ is the stacking of the vectors $\textbf{u}_k$ and then of the vectors $\textbf{v}_k$ for each of the $K$ modes in the Gaussian mixtures:
\begin{equation}\label{eq:fv-4}
\Phi(I) = [..., \textbf{u}_k, ..., \textbf{v}_k, ...]^{\top}.
\end{equation}

The improved Fisher Vector (IFV)~\cite{perronnin2010eccv} made improvements of the representation by two means, one is employing a non-linear additive kernel instead of the linear one, and the other is applying a normalization to the generated feature vectors.

\subsection{Color-based drawing style description}\label{sec:color}

\ Color is an another important visual stimuli that reflects the esthetics on a natural scene, as well as on a painting. Three types of local color descriptors are examined in this work, including CN, DD, and RCC.

{\it CN} - Color Names Descriptor~\cite{Weijer2009,Benavente2008}. Color names are linguistic labels when human perceives color in nature scenes, where a total of eleven basic color terms are defined, namely the black, blue, brown, grey, green, orange, pink, purple, red, white, and yellow~\cite{berlin1991basic}. Based on this idea, in \cite{Weijer2009}, the eleven color names were learnt from Google images automatically by partitioning the color space into eleven regions. Then, colors in a local region $R$ can be represented by an 11-dimensional vector:
\begin{equation}\label{eq:cn11d}
<p(n_1|R), p(n_2|R), ..., p(n_{11}|R)>,
\end{equation}
where $p(n_i|R)$=$\frac{1}{N}\sum p(n_i|f(x))$, $n_i$ denotes the $i$th color name, $N$ is the number of pixels in $R$, and $f(x)$ returns the color value at location $x$.

{\it DD} - Discriminative Color Descriptor~\cite{Khan13}. Different from CN, DD partitions the color space by using the information theory, where the color values are clustered into a number of $r$ categories by minimizing their mutual information in image representation. Similar with CN, DD examines the probability of an image region falling into each category, which leads to a $r$-dimensional vector:
\begin{equation}\label{eq:dd11d}
<DD_1, DD_2, ..., DD_r>.
\end{equation}
The resulted DD descriptor was reported to hold good photometric invariance and high discriminative power at $r$=25 and 50, and the increasing of $r$ over 50 did not bring meaningful improvement on the performance~\cite{Khan13}. Therefore, we will test the 25- and 50-dimensional DD in this work.

Note that, the above CN and DD are local color descriptors, which are usually used in a bag-of-word framework for image classification~\cite{Weijer2009,Khan13}.

{\it RCC} - Regional Color Co-occurrence Descriptor~\cite{zou2015icip}. Based on the observation that color-collocation patterns are more robust than color-distribution patterns in image and object recognition~\cite{zou2017tmm}, the RCC descriptors exploit the color collocations in the neighboring regions in the images, and encode the color collocation features using a regional color co-occurrence matrix.

Let $Z$ = $\{z_a|a=1,2...,n\}$ be a set of $n$ regions, where for
$\forall (a, b)$, $a\neq b$, $z_a \cap z_b= \emptyset$. Let $S$ =
$\{s_i|i=1,2...,N\}$ be the set of $N$ possible color features. For
each region $z_a$, the frequency of every color feature is calculated
and a color feature histogram $H^a$, $a=1,2,...,n$ can be computed by
\begin{equation}\label{eq:rcc-1a}
H^a =(H^a_1, H^a_2, ..., H^a_N),
\end{equation}
where $H^a_i$ is the number of occurrence of color feature $s_i$ in
the region $z_a$. Similarly, a color feature histogram $H^b$ can be constructed in region $z_b$,
\begin{equation}\label{eq:rcc-hb}
H^b =(H^b_1, H^b_2, ..., H^b_N),
\end{equation}
Then, the color co-occurrences between region $z_a$ and region $z_b$ can be computed by
\begin{equation}\label{eq:rcc-3}
\Phi ^{a,b} = \textbf{Norm}\{(H^a)^{\textsf{T}} \cdot H^b\},
\end{equation}
where $\textbf{Norm}\{\}$ denotes an $l$-1 normalization to an
$N\times N$ matrix. The summation of all $\Phi ^{a,b}$ over the whole image is taken as the final color descriptor.

\subsection{DCNN-based drawing style description}\label{sec:dcnn}

\ Machine learning has been advanced in the past several decades~\cite{Deng2015scis}. As an important machine learning technique, deep learning~\cite{science2006,Bengio06} has attracted wide attention since the beginning of this decade~\cite{zhouzh2016scis}. In visual cortex study, the
brain is found to have a deep architecture consisting of several layers. In visual activities, signals flow from one brain layer to the next to obtain different levels of abstraction. By simulating the function of the deep architecture of brain, deep learning uses a set of algorithms in machine learning to attempt to model high-level abstractions in data by using model architectures composed of multiple non-linear transformations learning~\cite{Deng2013trends}.

\begin{figure*}[t!]
    \centering
    \hspace{2mm}
    \includegraphics[width=0.9\linewidth]{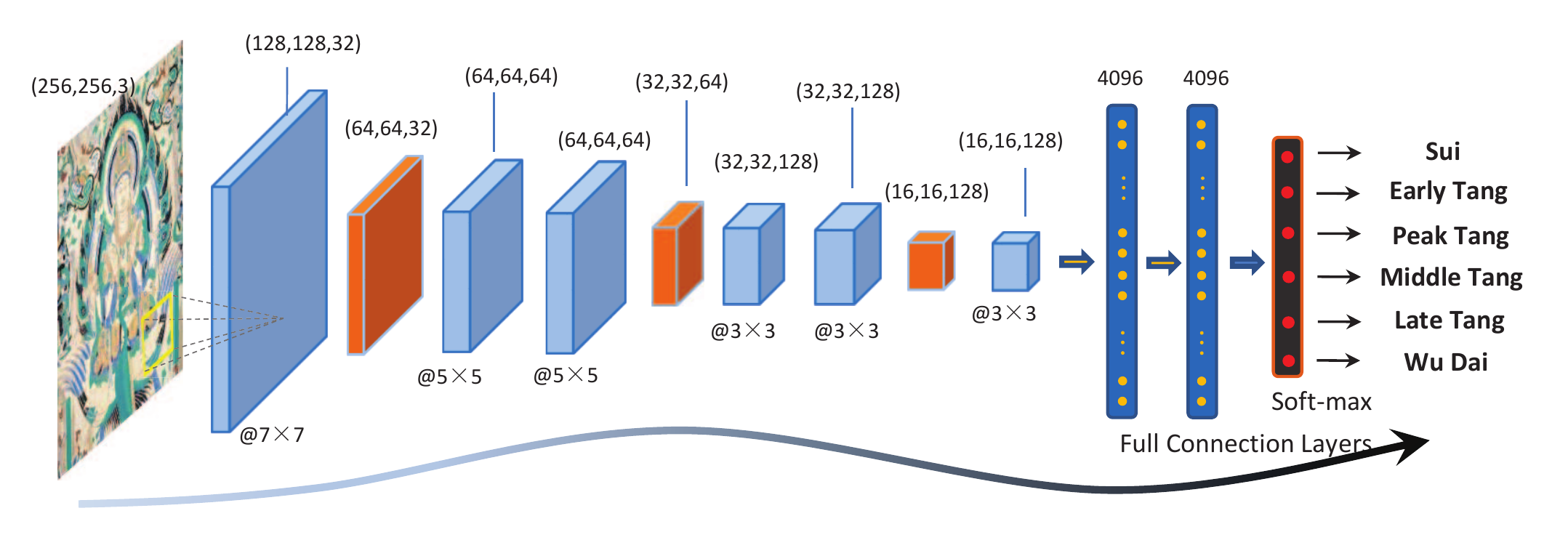}
    \caption{Architecture of the proposed DunNet neural network.}
    \label{fig:dcnn-arch}
\end{figure*}

Among various types of deep learning architectures, the deep convolution neural network (DCNN)~\cite{lecun1989backpropagation} was found to be the most effective one in handling images. A convolution neural network (CNN) is a type of feed-forward artificial neural network in which the connectivity pattern between its neurons is inspired by the organization of the animal visual cortex, whose individual neurons are arranged in such a way that they respond to overlapping regions tiling the visual field.~\cite{lecun1989backpropagation}. In recent year, with the development of advanced learning strategies and high-speed parallel computing technologies, deep-structured CNNs have been possible and the DCNN has attracted wide attention as it shows astounding improved performance in solving many computer vision problems. For image classification, a number of typical DCNN algorithms have been proposed, e.g., the AlexNet~\cite{krizhevsky2012imagenet}, the GoogLeNet~\cite{sc2015googlenet} and the VGGNet~\cite{simonyan2014very}, etc. In recent years, DCNN has also demonstrated powerful talents in painting style analysis~\cite{Gatys2016cvpr,Zisserman2014eccvw}. Inspired by the success of DCNN in image and painting information abstraction, we purposely design a DCNN-based classifier for Dunhuang mural painting classification, and employ the feature-leaning block of it as a third descriptor - DunNet - to construct the potential visual codes for drawing-style description.

In our design, the DCNN is equipped with six convolution layers, three max-pooling layers and two fully connected layers. Details of DunNet are illustrated by Fig.~\ref{fig:dcnn-arch}. The last layer, which is called the softmax layer, outputs probabilities of the predicted classes, which consists of six nodes indicating the Sui, Early Tang, Peak Tang, Middle Tang, Late Tang and Wu Dai, respectively. The ReLU (Rectified Linear Units~\cite{Hinton2010icml}) is employed as the activation function, and the loss function is built on the cross entropy error. To employ DunNet as a feature extractor, we take the second fully connected layer, which we name it the deeply learnt visual codes, as a feature vector for image representation.

\subsection{Creation era prediction}\label{sec:voting}
\ A well-trained classification model with high accuracy is supposed to have obtained and used the discriminative features from the training data. Generally, the discriminative features have minimal inter-class similarity and maximal intra-class similarity, which is a required property for predicting the labels of data samples coming from outside the training set. After training our classification model on our dynasty-specified painting dataset, we use it as a tool to produce the discriminative features, with which we predict the class labels of new-coming paintings. A voting scheme is presented to enhance the credibility of prediction.

Let $M$ be the number of samples cropped from the original painting image, $X_j$ (1$\leq$$j$$\leq$$M$) be the predicted label of the $j$th sample, $N$ be the number of classes, $\mathcal{V}_i$ (1$\leq$$i$$\leq$$N$) be the number of votes obtained on the $i$th class, then the final class label $\mathcal{L}$ can be computed by Eq.~\ref{eq:voting},
\begin{equation}\label{eq:voting}
\mathcal{L}= \arg
\max\limits_{i}\{{V}_i|i=1,2, ..., N\}
\end{equation}
where
\begin{equation}\label{eq:voting1}
\mathcal{V}_{i}=\sum\limits_{j=1:M} \xi^{(+,-)}(X_j, i),
\end{equation}
and
\begin{equation}\label{eq:voting2}
\xi^{+}(X_j, i)=\{1 |X_j = i\},\ \xi^{-}(X_j, i)=\{0 |X_j \neq i\}.
\end{equation}

Considering a painting has an unknown size, which is different from the samples in our training set, we first sample a number of images from the painting to be dated, then predict the era labels for all of them, and finally take the predicted class label as a vote, and assign the era gaining the most votes as a label to the original painting. To reduce the impact of scale variation, we sample the subimages with five different scales, which are from 0.8 to 1.2 at an interval of 0.1 to the size of data samples in the training set. Specifically, 20 subimages are randomly sample at each scale, and are resized to the size of the data samples in the training set for feature extraction and class label prediction. Hence, a total of 100 samples are utilized to predict the final era label of one input painting.

\subsection{Experimental settings}
\ For shape descriptor IFV{\scriptsize{sift}}, a dense sampling strategy is employed, in which multi-scale dense SIFTs are extracted from each input image, with a sampling step of 4, and a scale number of 5. For color descriptors CN and DD, features are extracted with a sampling step of 5, and a scale number of 2. Note that, CN produces 11-dimensional features, and is named as CN11. Two versions of DD are used, i.e., 25 dimensions and 50 dimensions, which are named as DD25 and DD50, respectively. For color descriptor RCC, a color codebook with 128 codes is computed using K-means clustering. In SIFT feature encoding, the improved fisher vector algorithm (IFV) equipped with a number of 128 Gaussian models is employed. While for CN and DD, a standard bag-of-words method is utilized~\cite{jyd2016scis,chenl2017asift}, using K-means clustering with a number of 512 centers. For DCNN, the training is equipped with a learning rate of 0.001, a decay rate of 0.5 under a step of 4,000 iterations, and a total of 50,000 iterations. The training dataset for DCNN is augmented into a number of 54,000, where image rotation is performed from -10 to 10 degrees under an interval of 2.5 degrees, and all images are horizontally flipped. In classification, support vector machine (SVM) is employed, using a $\chi^2$ kernel. In the multi-class classification, a one-vs-all strategy is used.

\section{Results}

\subsection{Mural painting classification}
\ In the classification, 3000 painting images in Dunhuang-E6 are used for training, and the remaining 700 are used for testing. Note that, the percentage of training/testing samples is identical for different class of the paintings. For shape-based drawing style description, IFV{\scriptsize{sift}} is employed. For color-based drawing style description, CN11, DD25, DD50, and RCC are used. For drawing style with unknown forms, the AlexNet, GoogLeNet and DunNet are used. The classification accuracies of these methods, as well as their combinations, have been shown in Table~1. The classification experiments are performed in two ways, one is the 6-class classification, where the results are listed in the last column in Table~1, and the other one is the binary classification, i.e., one-vs-one, where the results are listed in the 2nd to 6th columns in Table~1.

\renewcommand{\arraystretch}{1.1}
\begin{table*}[t!]
\center \caption{Classification accuracies of different methods and their combinations (\%)}\vspace{2mm}
\footnotesize
\begin{tabular}{ccccccc}
\hline
  Descriptor & \tabincell{c}{Sui - \\Early Tang} & \tabincell{c}{Early Tang - \\Peak Tang} & \tabincell{c}{Peak Tang - \\Middle Tang} & \tabincell{c}{Middle Tang - \\Late Tang} & \tabincell{c}{Late Tang - \\Wu Dai} & Six Dynasties\\
  \hline 
  IFV{\scriptsize{sift}} & 82.11 & 73.61 & 82.45 & 67.08 & 80.20 & {\textbf{59.90}}\\
  \hline
  CN11 & 76.94 & 68.30 & 75.64 & 69.88 & 75.66 & 46.02\\
  DD25 & 70.94 & 67.14 & 70.29 & 71.87 & 66.48 & 39.94\\
  DD50 & 70.16 & 69.10 & 75.23 & 72.05 & 65.86 & 43.78\\
  RCC & 84.77 & 75.26 & 83.57 & 70.73 & 78.83 & {\textbf{59.34}}\\
  \hline
  AlexNet(256) & 85.91 & 76.77 & 81.30 & 77.50 & 84.84 & 60.37\\
  GoogLeNet(256) & 86.51 & 77.44 & 83.23 & 80.12 & 81.88 & 60.83\\
  DunNet(256) & 88.11 & 81.03 & 84.40 & 75.65 & 83.03 & {\textbf{62.95}}\\
   \hline
  IFV{\scriptsize{sift}}\ +\ RCC & 83.86 & 82.46 & 91.98 & 72.51 & 87.57 & 69.60\\
  IFV{\scriptsize{sift}}\ +\ DunNet & 88.98 & 83.78 & 85.91 & 77.88 & 83.92 & 64.52\\
  IFV{\scriptsize{sift}}\ +\ RCC + DunNet & 88.51 & 86.07 & 91.62 & 77.25 & 89.40 & {\textbf{71.64}}\\
\hline
\end{tabular}\label{tbl:allmethods}
\end{table*}

\begin{figure*}[t!]
\centering
\begin{minipage}[b]{0.32\linewidth}
  \centering
  \centerline{\includegraphics[width=5.96cm]{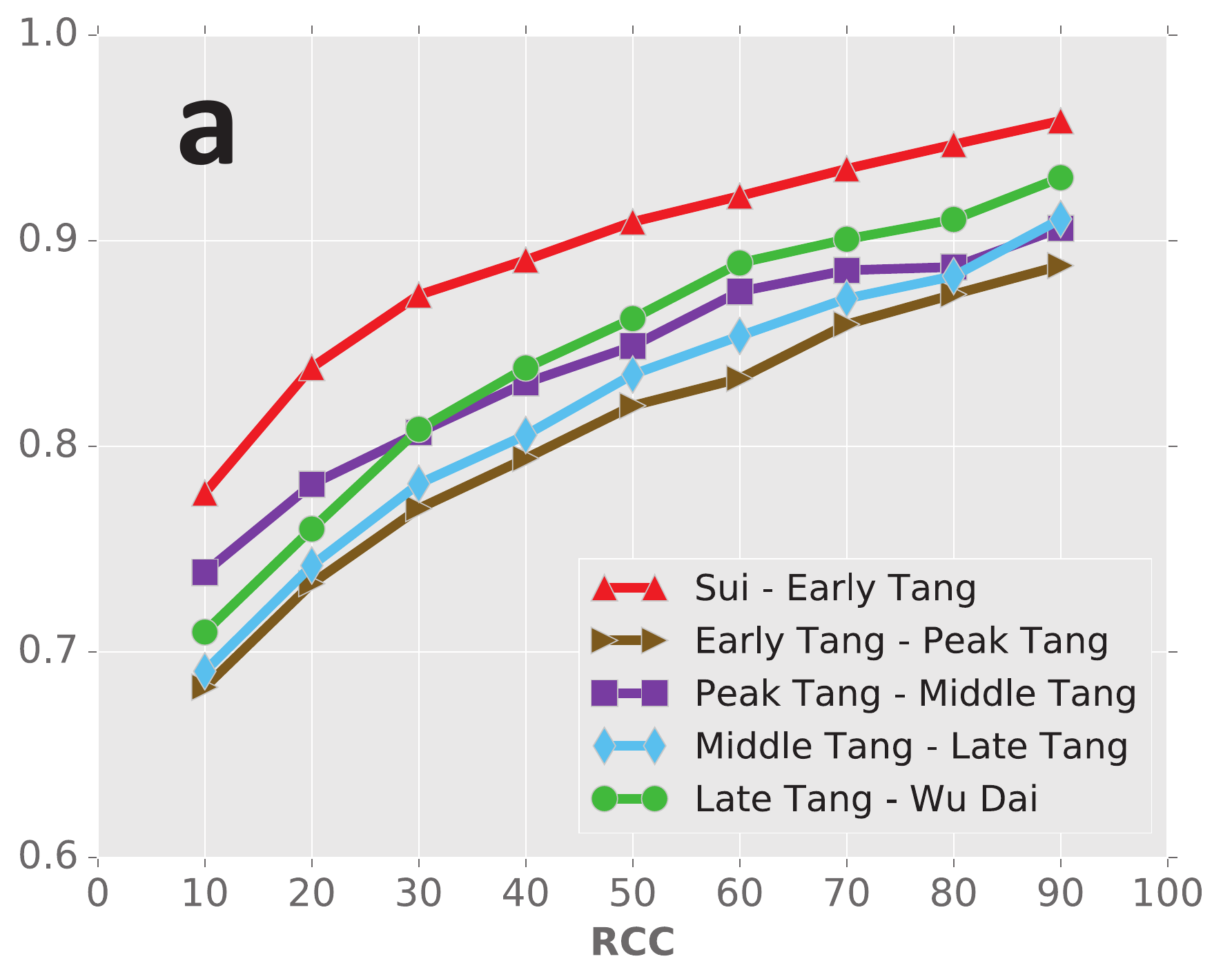}}
  \medskip
\end{minipage}
\begin{minipage}[b]{0.32\linewidth}
  \centering
  \centerline{\includegraphics[width=5.96cm]{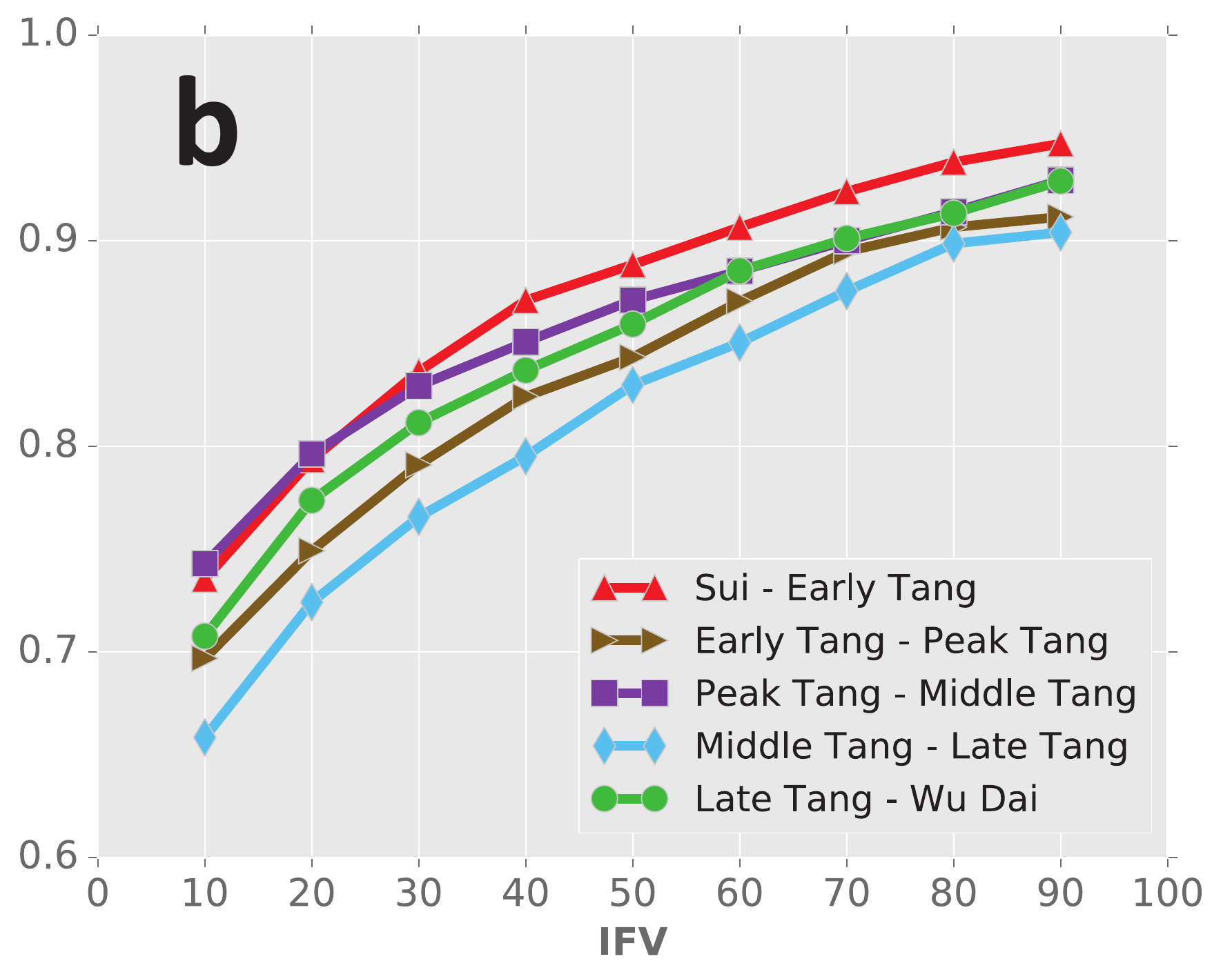}}
  \medskip
\end{minipage}
\begin{minipage}[b]{0.32\linewidth}
  \centering
  \centerline{\includegraphics[width=5.96cm]{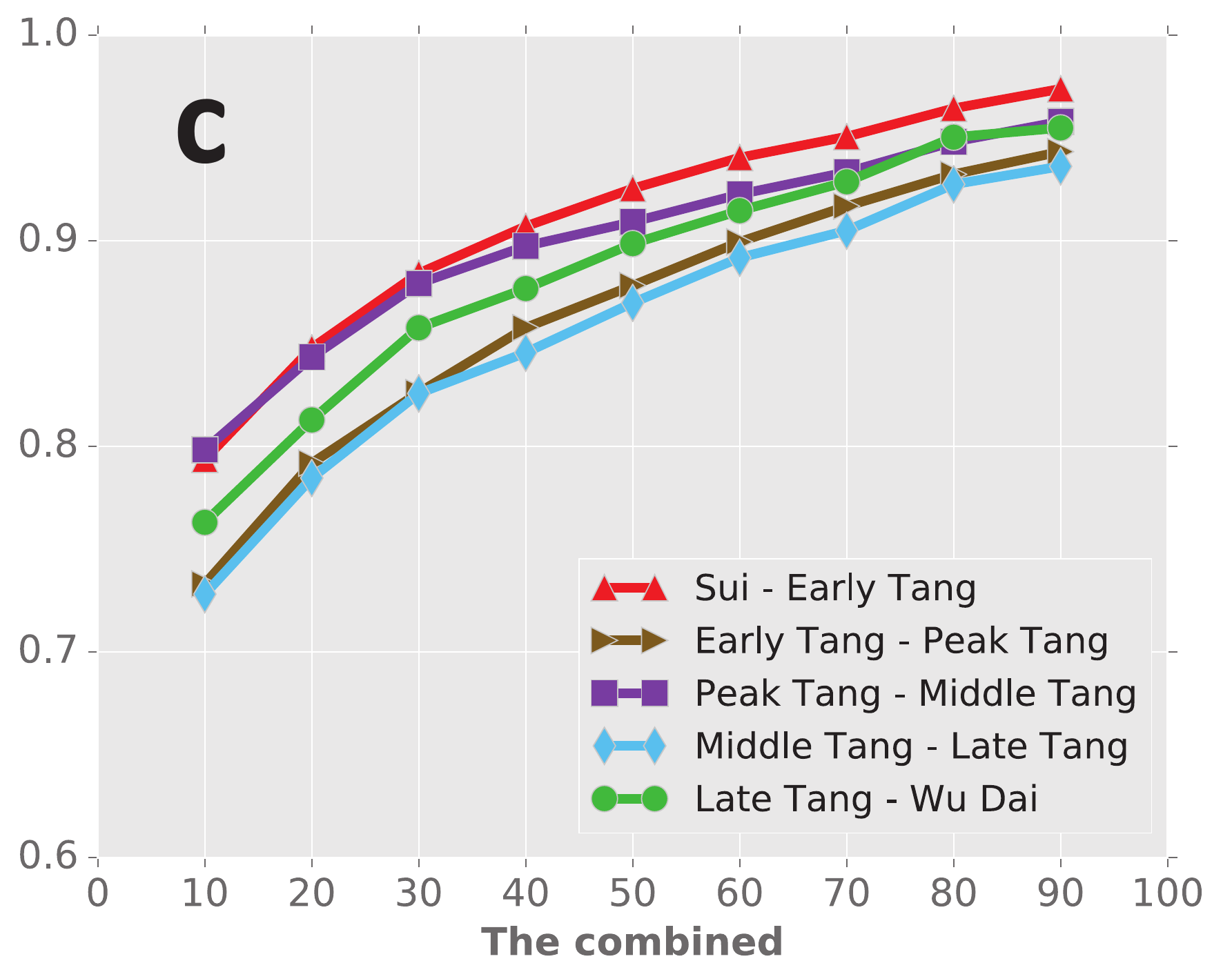}}
  \medskip
\end{minipage} \vspace{-3mm}
\caption{Classification accuracies obtained by binary classifiers (\%). (a) RCC. (b) IFV. (c) IFV+RCC+DunNet.} \label{fig:ratio}
\end{figure*}

For the 6-class classification, the values under the `Six Dynasties' show that, the RCC achieves the highest accuracy among the four color-based methods, which indicates that the color collocations encoded by RCC is a more discriminative pattern than the color distribution encoded by CN and DD in drawing style description. In fact, some other color descriptors such as the normalized color histogram\cite{Sande2010} and Hue histogram are also tested and found to have inferior performance. The shape-based method, IFV{\scriptsize{sift}}, obtains an accuracy of 59.90\%, which is slightly higher than that of RCC, which indicates that shape is also a strong indicator of drawing styles for mural paintings of Dunhuang. The proposed DunNet gains an accuracy 3\% higher than that of shape or color, which demonstrates the superiority of the deeply learned features over the traditional handicraft features in describing drawing style of Dunhuang paintings. AlexNet and GoogLeNet show slightly lower performance than DunNet. A possible reason would be that the limited training data may not be enough to fully train the more complex networks.

For the binary classification, five datasets are constructed for performance evaluation, where paintings of each pair of neighboring eras are combined into one dataset. With regard to the names of the neighboring eras, the five datasets are named as `Sui - Early Tang', `Early Tang - Peak Tang', `Peak Tang - Middle Tang', `Middle Tang - Late Tang', and `Late Tang - Wu Dai', respectively. By fixing the training part and the testing part, binary classifiers using color, shape, and deep-learning features, and their combinations are evaluated on the five datasets, and the results have been given in Table~1. We can see that, the combined method, i.e., IFV+RCC+DunNet, achieves accuracies higher than 86\% on all the datasets except for the `Middle Tang - Late Tang'. One possible reason is that the painting styles are so similar between Middle Tang and Late Tang that the color, shape and deep-learning features cannot capture their discriminations.

In addition, the binary classifiers are tested by varying the number of training samples from 10\% to 90\% at an interval of 10\% on the 3700 samples in Dunhuang-E6. Note that, at each test, the training samples are randomly selected from the 3700 images, and the rest are taken as the testing samples. The classification procedure is repeated 20 times at each test and the classification accuracies are averaged. The results are shown in Fig.~\ref{fig:ratio}. From Fig.~\ref{fig:ratio} we can see, the increasing of training samples leads to improved classification accuracy for all the classifiers. The classifiers achieve relatively higher performance on `Sui- Early Tang' and lower performance on `Early Tang - Peak Tang' and `Middle Tang - Late Tang' in terms of both color (RCC)- and shape (IFV)- based features. It simply demonstrates the level of difference of the drawing styles between different eras. Similar properties are also observed on the classifiers using the combined features, as shown in Fig.~\ref{fig:ratio}(c), and significantly improved performances have been achieved on all the five datasets.

\subsection{Dating mural paintings}
\begin{figure*}[htbp]
    \centering
    \includegraphics[width=0.9\linewidth]{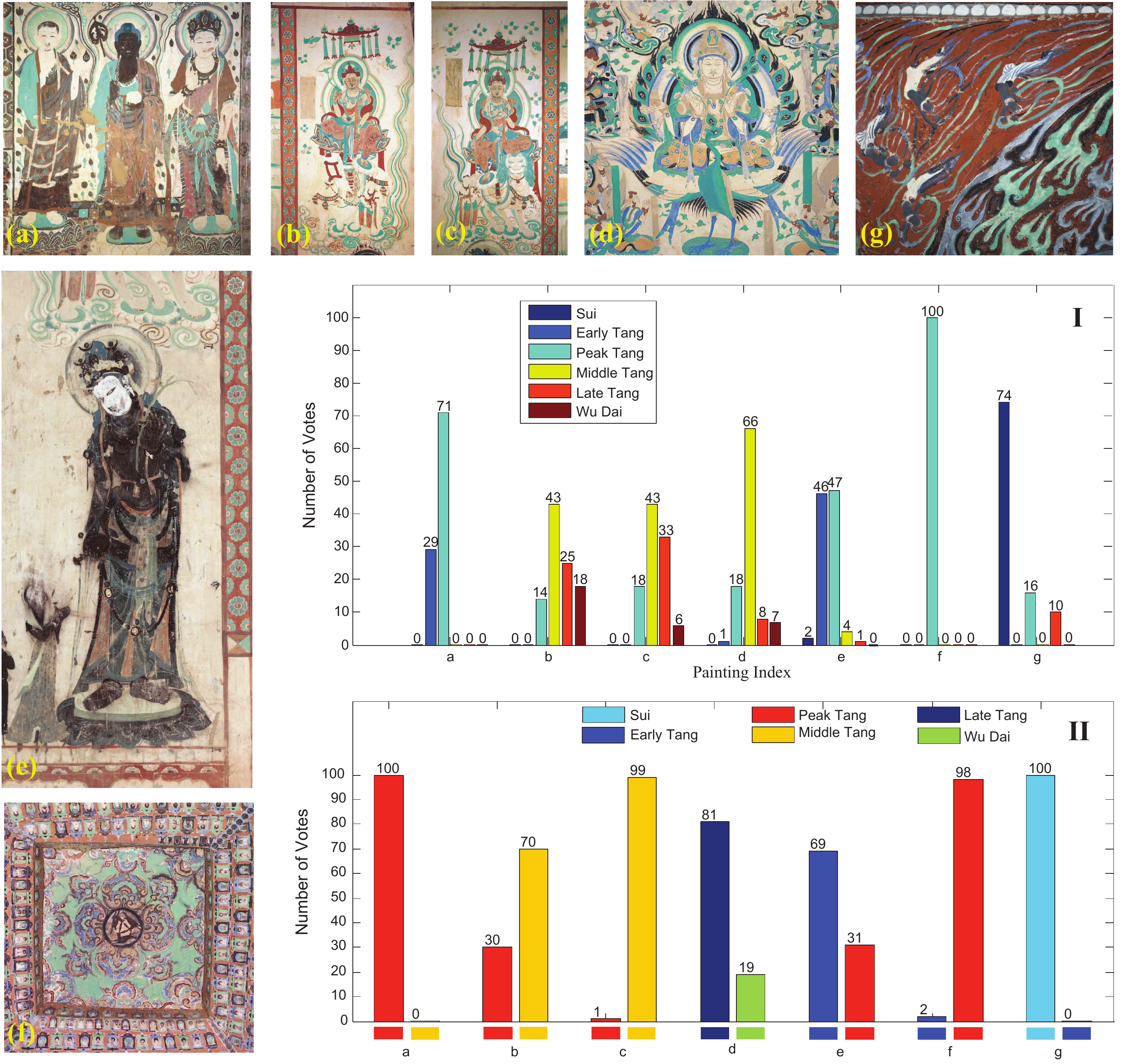}
    \caption{Seven mural paintings from Mogao Grottoes. Note that, (a)-(f) are from grotto $\texttt{\#}$205, and (g) is from grotto $\texttt{\#}$206. (a) A mural painting located on the south wall of grotto $\texttt{\#}$205. (b)~A mural painting on the south side of the west wall.
    (c)~A mural painting on the north side of the west wall. (d) A mural painting of Peacock King, on the aisle's ceiling of grotto $\texttt{\#}$205. (e) A painting of Guanyin Buddha on south side of the west wall of grotto $\texttt{\#}$205. (f) A painting from the central ceiling of grotto $\texttt{\#}$205. (g) A Flying-Apsaras painting from grotto $\texttt{\#}$206. The classification results are shown in the two bar chars. (I) Voting results on the seven paintings using the 6-class classifier. (II) Voting results of the seven painting images using binary classifiers.}
    \label{fig:seven700-7}
\end{figure*}

\ The above trained classification models can be applied to recognize the drawing styles of mural paintings, and further predict the creation eras of them. In this subsection, we perform two experiments to predict the creation era of seven paintings in DunHuang-P7, as shown in Fig.~\ref{fig:seven700-7}. For each painting in DunHuang-P7, 100 samples are cropped from the original painting image under five scales with 20 per scale, as describe in Section~\ref{sec:voting}. First, we directly apply the multi-class classifier - `IFV+RCC+DunNet' - trained on DunHuang-E7 to test the seven paintings. Each sample obtains a predicted era label, and the final results have been shown in Fig.~\ref{fig:seven700-7}(I). From Fig.~\ref{fig:seven700-7}(I) we can see, the paintings in Fig.~\ref{fig:seven700-7}(a), (b), (c) and (g) get the majority votes corresponding to the official era label given by Dunhuang experts. For painting in Fig.~\ref{fig:seven700-7}(e), there's no dominance in the voting, as the top one and top two vote receivers hold the number of votes that are very close to each other. However, the top one and top two rightly cover the two most possible eras as suggested by experts, i.e., Early Tang and Peak Tang. For the painting in Fig.~\ref{fig:seven700-7}(f), the dominant vote receiver is `Peak Tang', which fights against the alternative creation era, i.e.,  the previous official `Early Tang', and supports for the up-to-date official era label `Peak Tang'.

The above results provide some confidences to the classification-based era prediction, but it is not enough since the accuracy of the 6-class classification is 71.64\%. To make more convincing predictions, we use binary classifiers to recognize the drawing styles, as the binary classification models have high performance in discerning the creation era (through drawing style) of a painting once two candidate eras are given. For the seven paintings in DunHuang-P7, each painting has been officially assigned an era label by Dunhuang Research Academia, as marked in green in Table~2. Meanwhile, each painting has been given an alternative label of the creation era as marked in red, in case it was not painted in the era officially given. For the first three paintings in Fig.~\ref{fig:seven700-7}, i.e., (a)-(c), we would like to clarify if they were painted in Peak Tang or in Middle Tang. Accordingly, the binary classifier trained on the `Peak Tang - Middle Tang' dataset, with an accuracy of 91.62\%, is used to recognize the drawing style of them. The rest four paintings are also evaluated using the according binary classifiers. The results are plotted in Fig.~\ref{fig:seven700-7}(II). It can be seen from Fig.~\ref{fig:seven700-7}(II), each classification has achieved one single dominant vote receiver. Specifically, the voting results have made validation to the correctness of the official creation-era labels for paintings in Fig.~\ref{fig:seven700-7}(a), (b), (c) and (g). For the Peacock King painting in Fig.~\ref{fig:seven700-7}(d), the voting results imply the drawing style of Late Tang other than the Wu Dai. The voting results also support the up-to-date official dating results on the paintings in Fig.~\ref{fig:seven700-7} (e) and (f), which were previously dated as `Peak Tang' and `Early Tang', respectively.

\renewcommand{\arraystretch}{1.0}
\begin{table}[t!]
\small\center \caption{The possible creation era of the seven paintings in Fig.~\ref{fig:seven700-7}(a)-(g). The green denotes the up-to-date official creation era given by the Dunhuang Research Academia, and the red with question mark denotes the most probable creation era in alternative.}\vspace{2mm}
\begin{tabular}{C{0.8cm}C{0.8cm}C{0.8cm}C{0.8cm}C{0.8cm}C{0.8cm}C{0.8cm}}
\hline
  \tabincell{c}{Painting \\ Index} & {\ Sui \ } & \tabincell{c}{Early \\ Tang} & \tabincell{c}{Peak \\ Tang} & \tabincell{c}{{\footnotesize Middle} \\ Tang} & \tabincell{c}{Late \\ Tang} & \tabincell{c}{Wu \\ \ Dai\ } \\
 \hline 
  (a) &  &   &  \multicolumn{1}{>{\columncolor{green}}l}{} &  \multicolumn{1}{>{\columncolor{red}}l}{\ \ \ \ ?} &   &  \\
  \hline
  (b) &  &   & \multicolumn{1}{>{\columncolor{red}}l}{\ \ \ \  ?} &  \multicolumn{1}{>{\columncolor{green}}l}{} &   &  \\
    \hline
  (c) &  &   & \multicolumn{1}{>{\columncolor{red}}l}{\ \ \ \ ?} &  \multicolumn{1}{>{\columncolor{green}}l}{} &   &  \\
    \hline
  (d) &  &   &   &   &  \multicolumn{1}{>{\columncolor{red}}l}{\ \ \ \ ?} & \multicolumn{1}{>{\columncolor{green}}l}{} \\
    \hline
  (e) &  & \multicolumn{1}{>{\columncolor{green}}l}{}  & \multicolumn{1}{>{\columncolor{red}}l}{\ \ \ \ ?}  &   &   &  \\
    \hline
  (f) &  & \multicolumn{1}{>{\columncolor{red}}l}{\ \ \ \ ?}  & \multicolumn{1}{>{\columncolor{green}}l}{}  &   &   &  \\
    \hline
  (g) & \multicolumn{1}{>{\columncolor{green}}l}{} & \multicolumn{1}{>{\columncolor{red}}l}{\ \ \ \ ?}  &   &   &   &  \\ 
\hline
\end{tabular}\label{tbl:eras}
\end{table}

\section{Discussion}

Scientific dating is very important in archaeology~\cite{aitken2014book,pike2012science}. This work predicts the creation era of ancient mural paintings by examining their drawing styles. The creation-era dating problem is mapped into a drawing-style classification problem, and the combined shape, color, and deep-learning features are used to describe the drawing styles. Considering that the seven paintings with controversy in dating have potential creation eras spanning from Sui to Wu Dai, a dataset Dunhuang-E6 with 3860 painting images collected from Sui to Wu Dai is constructed and used to train the classification model. The high performances in the binary classification demonstrate that the combined color, shape and deep-learning features are capable of describing and discriminating the drawing styles of Dunhuang paintings in the selected six different eras.

For Fig.~\ref{fig:seven700-7}(a), (b), (c) and (g), they are four paintings with least controversy in creation-era dating. The classification results in Fig.~\ref{fig:seven700-7}(I) and (II) have all validated the correctness of the previous dating results. In fact, Fig.~\ref{fig:seven700-7}(a) is commonly dated into Peak Tang, while Fig.~\ref{fig:seven700-7}(b) and (c) are dated into Middle Tang. It supports the assumptions that the painting job in grotto $\texttt{\#}$205 was handicapped by the wars of Anshi Rebellion, which was a division of the Peak Tang and Middle Tang. For Fig.~\ref{fig:seven700-7}(g), though the grotto $\texttt{\#}$206 was caved in the Early Tang's region, the computing method concludes this painting is actually in Sui's drawing style.

For painting in Fig.~\ref{fig:seven700-7}(e), the controversy in its creation era lies in dating it into Early Tang or Peak Tang. Wen-Jie Duan, the highly respected former President of DunHuang Research Academia, dated this painting into Peak Tang, while Hui-Min Wang, a distinguished expert in Archaeology Institute of Dunhuang Research Academia, dated it into Early Tang~\cite{wanghm2010}. The classification results in Fig.~\ref{fig:seven700-7}(I) show that the `Early Tang' and `Peak Tang' are the two main vote receiver, and their numbers of votes are very close, which indicates the difficulty in classifying it into Peak Tang or Early Tang. However, when we use a binary classifier, which is trained on the `Early Tang - Peak Tang' dataset, the results indicate the painting is in an Early Tang's drawing style, as shown in Fig.~\ref{fig:seven700-7}(II). While for the painting in Fig.~\ref{fig:seven700-7}(f), the dominant vote receiver is `Peak Tang' both in the multiclass case and in the binary case, which strongly points to the controversy in the creation era - the Peak Tang against the Early Tang. Although the era labels for Fig.~\ref{fig:seven700-7} (e) and (f) have been officially updated as `Early Tang' and `Peak Tang', respectively, there are still some doubles. The dating results in our study have been a strong support to Hui-Min Wang's dating results, and validated the correctness of the up-to-date official era label.

For the paining `Peacock King' in Fig.~\ref{fig:seven700-7}(d), the official-dated era is Wu Dai. In the binary classification, we apply the binary classifier trained on `Late Tang - Wu Dai' dataset to do the classification. The result in Fig.~\ref{fig:seven700-7}(II) points out it is in Late Tang's style rather than the Wu Dai's style. While in the multiclass case, the result in Fig.~\ref{fig:seven700-7}(I) points out it is in Middle Tang's style (with one sole dominant vote receiver). Both results indicate a signal that the paining `Peacock King' may be not in Wu Dai's style. It brings a valuable new controversy to the Dunhuang research community for further investigation.

\section{Conclusion}\label{sec:conc}
In this study, the problem of mural-painting dating was formulated as a problem of drawing-style classification. Several sophisticated color and shape descriptors and one DCNN-based descriptor were used to describe the drawing styles. A number of 3860 mural paintings of 6 eras were collected from 194 grottoes. The proposed dating method was constructed by training a classification model on these era-determined paintings. Experimental results showed that the DCNN features outperformed traditional handicraft features in drawing style classification, and the combined features achieved higher classification accuracies. The proposed dating method was applied to seven mural paintings of two grottoes which were previously dated with controversies. Among the seven new dating results, six were approved by the expert of Dunhuang Research Academia and were consistent with the up-to-date official era labels, and one was counted as bringing up a new research topic to the community. The proposed method has in the first time provided a scientific and qualitative manner to support the era prediction and reduce the uncertainties in dating ancient paintings.

\section*{Acknowledgements}

This work was supported by National Basic Research Program of China (Grant No.
2012CB725303), Major Program of Key Research Institute on Humanities and Social Science of the Chinese
Ministry of Education (Grant No. 16JJD870002), and National Natural Science Foundation of China (Grant
Nos. 91546106, 61301277). The authors would like to thank the Dunhuang Research Academia for providing the
mural paintings of Dunhuang-P7, and thank Mr.~Hui-Min WANG for helpful suggestions and discussions.

\flushend

\end{document}